\documentclass{edm_article}

\usepackage{graphicx}
\usepackage{xcolor}
\usepackage{natbib}
\usepackage{amsmath}
\usepackage{amssymb}
\usepackage{tikz}
\usepackage{booktabs}
\usepackage[most]{tcolorbox}
\usepackage[inline]{enumitem}
\usepackage{url} 
\usepackage{algorithmic}
\usepackage{algorithm}

\usepackage{tcolorbox}
\usepackage{colortbl}
\usepackage{pifont}  
\usepackage{array}
\usepackage{pgfplots}

\tcbuselibrary{skins,breakable}

\begin{document}

\title{Confusion-Aware Rubric Optimization for LLM-based Automated Grading}


%
%
%
%

\numberofauthors{7} 
\author{
\alignauthor
Yucheng Chu \\
       \affaddr{Michigan State University}\\
       \email{chuyuch2@msu.edu}
\alignauthor
Hang Li\\
       \affaddr{Michigan State University}\\
       \email{lihang4@msu.edu}
\alignauthor Kaiqi Yang\\
        \affaddr{Michigan State University}\\
       \email{kqyang@msu.edu}
\and  
\alignauthor Yasemin Copur-Gencturk\\
       \affaddr{University of Southern California}\\
       \email{copurgen@usc.edu}
\alignauthor Joseph Krajcik\\
       \affaddr{Michigan State University}\\
       \email{krajcik@msu.edu}
\alignauthor Namsoo Shin\\
       \affaddr{Michigan State University}\\
       \email{namsoo@msu.edu}
\and 
\alignauthor Jiliang Tang\\
       \affaddr{Michigan State University}\\
       \email{tangjili@msu.edu}
}

\maketitle

\begin{abstract}
Accurate and unambiguous guidelines are critical for large language model (LLM) based graders, yet manually crafting these prompts is often sub-optimal as LLMs can misinterpret expert guidelines or lack necessary domain specificity. Consequently, the field has moved toward automated prompt optimization to refine grading guidelines without the burden of manual trial and error. However, existing frameworks typically aggregate independent and unstructured error samples into a single update step, resulting in ``rule dilution'' where conflicting constraints weaken the model's grading logic. To address these limitations, we introduce \textbf{C}onfusion-\textbf{A}ware \textbf{R}ubric \textbf{O}ptimization (\textbf{CARO}), a novel framework that enhances accuracy and computational efficiency by structurally separating error signals. CARO leverages the confusion matrix to decompose monolithic error signals into distinct modes, allowing for the diagnosis and repair of specific misclassification patterns individually. By synthesizing targeted ``fixing patches" for dominant error modes and employing a diversity-aware selection mechanism, the framework prevents guidance conflict and eliminates the need for resource-heavy nested refinement loops. Empirical evaluations on teacher education and STEM datasets demonstrate that CARO significantly outperforms existing SOTA methods. 
These results suggest that replacing mixed-error aggregation with surgical, mode-specific repair yields robust improvements in automated assessment scalability and precision.

\end{abstract}


\section{Introduction}

The rapid shift toward personalized and competency-based learning has placed unprecedented challenges for traditional assessment frameworks. As personalized learning relies on understanding the specific needs for each learner, it necessitates high-quality formative assessment. In modern pedagogy, providing frequent and high-quality assessment feedback on students’ open-ended work is essential for diagnosing misconceptions and fostering deep conceptual understanding \cite{black1998assessment,hattie2007power,shute2008focus}. Yet these benefits are hard to deliver at scale. Constructed-response tasks, such as short answers or explanations, capture students’ reasoning more directly than multiple-choice formats, but they are substantially more time-consuming to assign and evaluate. Furthermore, they can exhibit nontrivial scorer disagreement even under rubric-based evaluation \cite{livingston2009constructed}. Notably, this disagreement often stems from the linguistic diversity and variability inherent in student answers. 
This ``grading bottleneck'' is reflected in teachers’ reported workloads, where scoring and correcting student work consumes a sizable share of working time across education systems \cite{oecd2025results}. As a result, educational practice often trades off authenticity for scalability by reducing the use of open-ended items or returning feedback after long delays. These compromises weaken the instructional value of assessment and motivate automated grading as a way to preserve the pedagogical benefits of constructed responses without prohibitive cost or latency \cite{attali2006automated,leacock2003c}.

Traditional methods of automated grading primarily relied on hand-crafted features, such as keyword matching, latent semantic analysis (LSA), or surface-level linguistic metrics \cite{foltz1999intelligent}. These approaches requires task-specific feature engineering and fails to capture the semantic nuances of student reasoning, resulting in rigid evaluations that often penalize creative but correct answers \cite{dikli2006overview}. The advent of deep learning introduced fine-tuning techniques using encoder-based models like BERT, which significantly improved semantic understanding and scoring accuracy \cite{devlin2019bert,mayfield2020should}. However, these models require extensive labeled datasets for training before applicable to the new tasks, making them impractical for dynamic classroom environments where assignments change frequently. 
Recent advancements demonstrate that large language models (LLMs) overcome this data barrier. Unlike their predecessors, LLMs can leverage zero-shot and few-shot capabilities (i.e., they can operate effectively with little to no labeled data) to grade open-ended student responses with performance approaching human-to-human agreement levels \cite{chen2025grading}. 
Despite this promise, the transition from human grading to LLM-driven assessment is fraught with technical hurdles. In particular, reliability is heavily contingent on the quality of the grading prompt, with the instructions and rubric operationalization constraining the model’s decision-making \cite{chen2025grading, jiang2024short}. Simply treating a human rubric as an instruction for an LLM is often sub-optimal; human rubrics, while intuitive for expert educators, often contain linguistic ambiguities that misalign with the causal reasoning of an LLM. This ``alignment gap'' necessitates extensive prompt engineering, a process that is often manual and driven by iterative trial and error, making it difficult to scale \cite{chu2025llm,zhang2025prompt}. Consequently, the field has moved toward \emph{automated prompt optimization} (APO), where algorithms iteratively refine instructions based on quantitative performance feedback \cite{zhou2022large,chu2025llm}.

While state-of-the-art frameworks like GradeOpt \cite{chu2025llm} leverage reflective optimization to diagnose errors, they are limited by their naive treatment of the error signal, which incurs substantial computational overhead when searching over large prompt spaces. Specifically, these methods aggregate heterogeneous error samples into a composite feedback signal, combining distinct failure types such as hallucination, excessive strictness, and rubric misunderstanding into a single optimization step without separation. This confounding of error types leads to \textit{rule dilution}. That is, by attempting to address a mixture of conflicting signals simultaneously, the optimizer generates generalized, weak instructions that fail to enforce clear decision boundaries. Consequently, to find even a marginally improved prompt, these methods rely on resource-heavy exploration, generating and evaluating large pools of candidates.

To address this, we introduce \textbf{CARO} (\textbf{C}onfusion-\textbf{A}ware \textbf{R}ubric for \textbf{O}ptimization), a framework that achieves superior accuracy and computational efficiency by structurally separating error signals. We observe that grading errors are not random mixtures, but structured clusters. CARO leverages the confusion matrix to decompose the monolithic error signal into distinct \textbf{error modes}, eliminating the inefficiency of trying to address conflicting error types simultaneously. Specifically, CARO targets the single \textbf{dominant error mode} at each step (e.g., a specific tendency to confuse a score of 0 with 1), transforming the optimization into a structured \textbf{diagnosis-and-repair} pipeline. By isolating a single, clear failure pattern, CARO generates targeted ``rule patches'' that are specific and highly effective. Contributing to its precision, CARO produces robust prompts with significantly fewer iterations and lower token usages compared to the existing APO baselines. Our empirical results confirm that CARO significantly outperforms these frameworks across multiple datasets while utilizing the same cost-effective backbone (\texttt{GPT-4o-mini}\footnote{\url{https://platform.openai.com/docs/models/gpt-4o-mini}}).

\section{Related Work}
\label{sec:related}

\paragraph{Automated short-answer grading (ASAG)} Early ASAG systems relied on hand-crafted lexical features and latent semantic analysis to predict scores \citep{mohler2009text}. The field later transitioned to neural architectures, utilizing recurrent models and Transformer encoders trained on graded corpora to capture deeper semantic dependencies \citep{vaswani2017attention, haller2022survey}. Despite their improved accuracy, these supervised pipelines require extensive labeled data for each specific question and lack the interpretability of rubric-based assessment. This limitation motivated the shift toward Large Language Models (LLMs), which offer zero-shot flexibility and the capacity to generate human-readable feedback alongside scores.

\paragraph{LLM-based grading and prompt optimization} Recent studies demonstrate that LLMs can achieve high agreement with human graders, though performance remains heavily dependent on rubric specificity and prompt design \citep{jiang2024short, chamieh2024llms}. While frameworks like RAG can improve domain grounding \citep{chu2025enhancing}, the challenge of manual tuning has accelerated the adoption of Automatic Prompt Optimization (APO). APO treats instructions as trainable parameters, ranging from iterative meta-optimization in OPRO and DSPy \citep{yang2023large, khattab2024dspy} to the ``natural-language gradients'' used in ProTeGi \citep{pryzant2023automatic}. In the educational domain, GradeOpt applies these principles to refine grading guidelines via self-reflection \citep{chu2025llm}. However, these systems generally aggregate heterogeneous errors into a single feedback signal, often obscuring the distinct failure patterns critical for precise assessment.

\begin{figure*}[htbp]
    \centering
    \includegraphics[width=\linewidth]{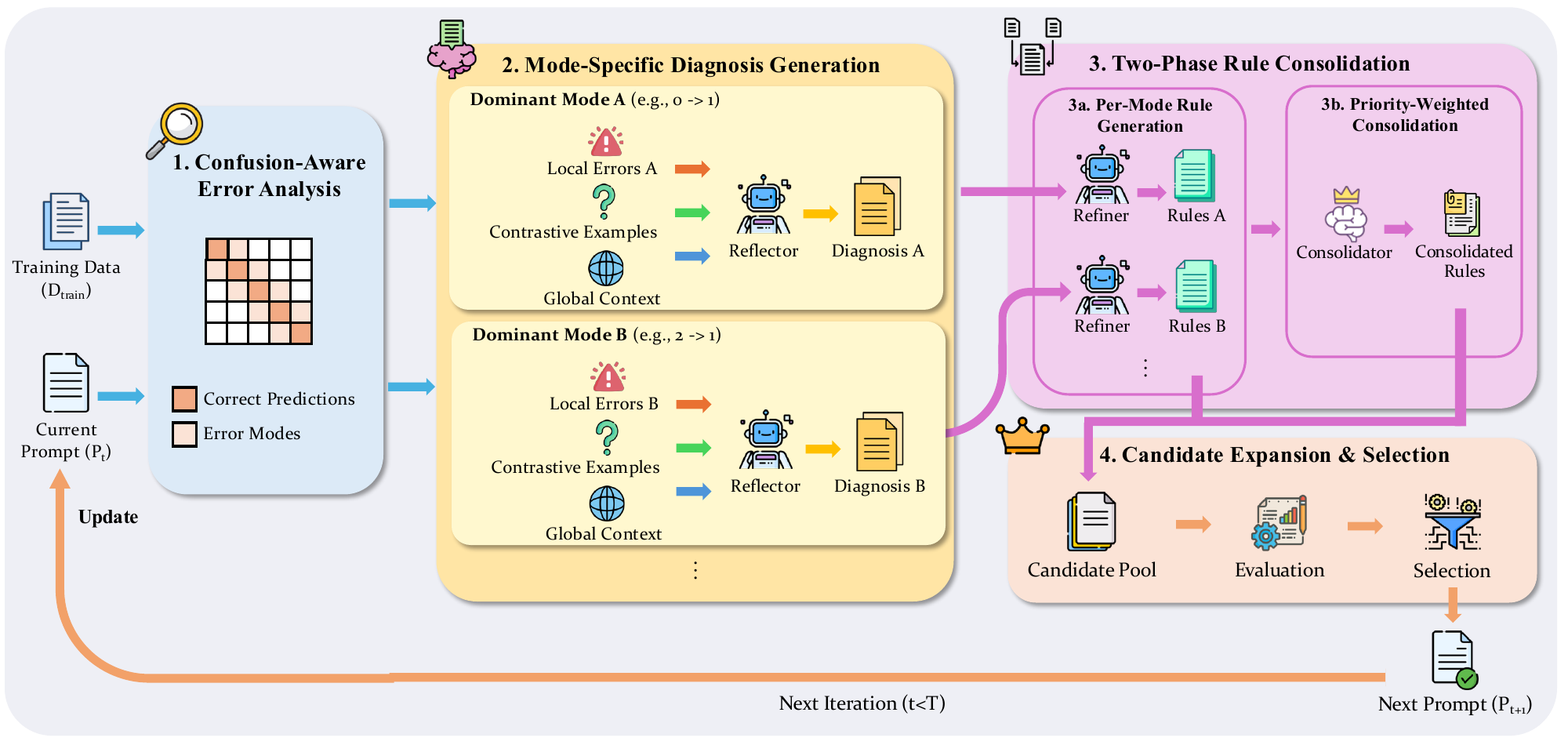}
    \caption{Framework of the CARO algorithm.}
    \label{fig:framework}
\end{figure*}

\section{Problem Formulation}
We formalize the task of automated rubric optimization as a discrete search problem over the space of natural language instructions. Unlike fine-tuning LLM parameters, which requires extensive labeled data and computational resources, we focus on optimizing the rubric prompt in the text space. This approach leverages the in-context learning capability of pre-trained LLMs while maintaining interpretability so that graders can inspect and validate the optimized rubric before deployment.
Let $\mathcal{D} = \{(x_n, y_n)\}_{n=1}^N$ be a dataset of student responses $x$ and ground-truth scores $y \in \mathcal{Y}$, where $\mathcal{Y} = \{0, 1, \dots, K-1\}$ represents an ordinal grading scale. An LLM-based grader can be defined as a function $f_\theta$ that estimates the probability $P_\theta(y | x, \mathcal{P})$ of a score $y$ given a response $x$ and a rubric prompt $\mathcal{P}$. The predicted score is obtained as $\hat{y} = \arg\max_{y \in \mathcal{Y}} P_\theta(y | x, \mathcal{P})$. The objective of \textit{Automated Prompt Optimization (APO)} is to identify an optimal prompt $\mathcal{P}^*$ that maximizes a performance metric $\mathcal{M}$ (e.g., classification accuracy or quadratic weighted kappa) over a validation dataset:
\begin{equation}
    \mathcal{P}^* = \arg\max_{\mathcal{P} \in \mathcal{S}} \mathcal{M}\big(\{\hat{y}_n, y_n\}_{n=1}^N\big)
\end{equation}
where $\mathcal{S}$ denotes the discrete search space of candidate prompts. Since the optimization targets the natural language prompt directly to ensure the rubric remains human-readable, standard differentiable optimization methods applicable to continuous parameters are not viable. Instead, APO methods approximate the gradient by generating textual feedback $g$ based on the discrepancy between predicted scores $\hat{y}$ and true scores $y$, effectively performing a semantic descent step $\mathcal{P}_{t+1} \leftarrow \text{Update}(\mathcal{P}_t, g)$.

The goal of our work is to address the structural inefficiency problem of the feedback signal $g$ when generated from aggregated error data. In the context of ordinal grading, errors are not uniform; they constitute distinct directional error modes captured by the confusion matrix $\mathbf{C}$, where an entry $C_{ij}$ represents the frequency of true class $i$ being misclassified as $j$. Existing APO frameworks typically aggregate a batch of heterogeneous error samples by mixing instances of $i \to j$ confusion with $k \to l$ confusion to generate a single, composite feedback signal. We define this aggregation as the source of \textit{rule dilution}. 
By exposing multiple error modes simultaneously, such methods require the optimizer model to solve a multi-objective task that demands high reasoning capacity to disentangle competing patterns. To address this challenge, we adopt a multi-agent decomposition strategy that separates the complex task into focused sub-problems: (1) a \textit{Mode-Specific Reflector} that analyzes each confusion cell $(i,j)$ independently to generate targeted feedback, and (2) a \textit{Consolidator} that synthesizes per-mode refinements into a coherent, non-redundant rule set while maintaining global consistency.

Furthermore, this optimization landscape is characterized by high inter-dependency between grading criteria, leading to \textit{mode interference}. Addressing a specific error mode $C_{ij}$ in isolation can act as a perturbation to the decision boundary between classes $i$ and $j$. Due to the semantic continuity of the rubric, this perturbation often propagates to adjacent boundaries, inadvertently increasing the error rate in a connected mode $C_{jk}$ or $C_{ji}$. This phenomenon creates a non-monotonic optimization trajectory where the reduction of one error type systematically amplifies another. Consequently, the optimization objective shifts from merely minimizing the total error count to identifying a structured sequence of rule refinements that effectively \emph{decomposes} the confusion matrix $\mathbf{C}$ without causing regression in previously stabilized modes. Our framework addresses this via a decomposition strategy that first generates mode-specific rules with global context awareness, then consolidating them with explicit priority ordering and cross-mode safety checks.

\section{Method}

We propose CARO (\textbf{C}onfusion-\textbf{A}ware \textbf{R}ubric for \textbf{O}ptimization), a framework designed to overcome the limitations of aggregating feedback in automated rubric optimization. Existing methods often suffer from rule dilution, where conflicting error signals lead to vague instructions. To address this, CARO structurally decomposes the optimization problem using confusion matrices, treating grading errors not as random noise but as distinct, diagnosable modes. Our approach operates in three iterative phases: confusion-aware error analysis, mode-specific gradient generation, and priority-weighted rule consolidation. 

\subsection{Framework Overview}

Figure~\ref{fig:framework} illustrates the complete CARO pipeline. To efficiently navigate the extensive search space of natural language prompts, CARO employs an iterative, minibatch-based optimization loop. In each iteration $t$, rather than evaluating on the entire training set, which is computationally expensive and may obscure minority error patterns, we sample a focused subset of data (a \textit{minibatch}) to estimate the current rubric's performance. This iterative design allows the framework to progressively refine the rubric, addressing the most prominent error modes in early rounds before fine-tuning for edge cases.

The framework operates as a closed-loop optimization system with four interconnected components. \textbf{Components 1--3} form the core optimization loop within a single round: (1) Confusion-Aware Error Analysis grades the minibatch and extracts structured error patterns; (2) Mode-Specific Gradient Generation produces targeted feedback for each confusion cell using a Reflector LLM; and (3) Two-Phase Rule Consolidation synthesizes per-mode rules into a unified rubric. These three components pass information sequentially within each optimization round. 
\textbf{Component 4} (Candidate Selection and Sampling) operates at a different granularity as it governs the \textit{inter-round} dynamics. It acts as a bridge between iterations by determining which candidate prompts survive to round $t+1$ and constructing the specific training minibatch that will be used to evaluate it. This separation reflects the algorithmic structure: Components 1--3 generate candidate prompts within a single round, while Component 4 decides \textit{which} candidates propagate across rounds and \textit{what data} they will be evaluated on next. The dashed connection in the figure indicates this temporal decoupling.

The complete procedure is summarized in Algorithm~\ref{alg:caro}. Given an initial rubric prompt $\mathcal{P}_0$, a training dataset $\mathcal{D}_{\text{train}}$, and a LLM predictor $f_\theta$, our objective is to iteratively optimize the prompt to minimize classification errors while ensuring stability across different score categories. The optimization process functions as a directed search. At each iteration $t$, we first evaluate the current prompt on a minibatch to generate a confusion matrix $\mathbf{C}^{(t)}$ (Algorithm~\ref{alg:caro}, lines 4--7). From this matrix, we identify the top-$K$ error modes, denoted as $\mathcal{M} = \{(i_1, j_1), \dots, (i_K, j_K)\}$, ranked by frequency (line 8). Unlike standard methods that aggregate these errors, CARO generates distinct gradients for each mode (lines 10--16), which are subsequently consolidated into coherent rule updates (lines 18--23). This cycle repeats until convergence or a maximum step count $T$ is reached.

\begin{algorithm}[t]
\caption{CARO: Confusion-Aware Rubric Optimization}
\label{alg:caro}
\begin{algorithmic}[1]
\REQUIRE Initial rubric prompt $\mathcal{P}_0$, training data $\mathcal{D}_{\text{train}}$, LLM predictor $f_\theta$, rounds $T$, beam size $B$, top-$K$ modes, diversity weight $\lambda$
\ENSURE Optimized rubric prompt $\mathcal{P}^*$

\STATE $\mathcal{P}^{(0)} \gets \{\mathcal{P}_0\}$ \COMMENT{Initialize candidate set}
\FOR{$t = 1$ \TO $T$}
    \STATE \textcolor{gray}{\texttt{// 1. Confusion-Aware Error Analysis}}
    \STATE $\text{Minibatch}^{(t)} \gets \text{SBERTSample}(\text{TopMisconf}^{(t-1)}, \mathcal{D}_{\text{train}})$
    \FOR{each $\mathcal{P} \in \mathcal{P}^{(t-1)}$}
        \STATE Evaluate $\mathcal{P}$ on $\text{Minibatch}^{(t)}$ to obtain $\mathbf{C}^{(t)}_{\mathcal{P}}$
    \ENDFOR
    \STATE $\mathcal{M} \gets \text{TopKModes}(\mathbf{C}^{(t)}, K)$ \COMMENT{Identify top-$K$ error modes}
    
    \STATE \textcolor{gray}{\texttt{// 2. Mode-Specific Diagnosis Generation}}
    \STATE $\mathcal{G} \gets \emptyset$
    \FOR{each mode $(i, j) \in \mathcal{M}$}
        \STATE $\mathcal{E}_{i \to j} \gets$ Extract error examples with true label $i$, predicted $j$
        \STATE $\mathcal{E}^{+}_{i}, \mathcal{E}^{+}_{j} \gets$ Sample high-misconfidence correct examples
        \STATE $g_{i \to j} \gets \text{Reflector}(\text{Rules}^{(t)}, \mathcal{E}_{i \to j}, \mathcal{E}^{+}_{i}, \mathcal{E}^{+}_{j}, \mathbf{C}^{(t)}, \mathcal{M})$
        \STATE $\mathcal{G} \gets \mathcal{G} \cup \{g_{i \to j}\}$
    \ENDFOR
    
    \STATE \textcolor{gray}{\texttt{// 3. Two-Phase Rule Consolidation}}
    \STATE $\mathcal{R} \gets \emptyset$
    \FOR{each $(i, j) \in \mathcal{M}$}
        \STATE $r_{i \to j} \gets \text{Refiner}(\text{Rules}^{(t)}, g_{i \to j}, \mathcal{M} \setminus \{(i,j)\}, \mathbf{C}^{(t)})$
        \STATE $\mathcal{R} \gets \mathcal{R} \cup \{(r_{i \to j}, (i,j))\}$
    \ENDFOR
    \STATE $r_{\text{consolidated}} \gets \text{PriorityConsolidate}(\mathcal{R}, \mathbf{C}^{(t)})$
    
    \STATE \textcolor{gray}{\texttt{// 4. Candidate Expansion and Selection}}
    \STATE $\mathcal{P}_{\text{new}} \gets \{\mathcal{P} \oplus r : \mathcal{P} \in \mathcal{P}^{(t-1)}, r \in \mathcal{R} \cup \{r_{\text{consolidated}}\}\}$
    \STATE Score each $\mathcal{P} \in \mathcal{P}_{\text{new}}$ via UCB-based evaluation
    \STATE $\mathcal{P}^{(t)} \gets \text{DiverseSelect}(\mathcal{P}_{\text{new}}, B, \lambda)$ 
    \COMMENT{Eq.~\ref{eq:select}}
\ENDFOR
\STATE \textbf{return} $\arg\max_{\mathcal{P} \in \mathcal{P}^{(T)}} \kappa(\mathcal{P}, \mathcal{D}_{\text{val}})$
\end{algorithmic}
\end{algorithm}

\begin{figure*}[t]
\centering
\small
\begin{tcolorbox}[
    colback=white,
    colframe=black!70,
    title={\textbf{Reflector Prompt for Mode $(0 \to 1)$ Confusion}},
    fonttitle=\bfseries\small,
    boxrule=0.5pt,
    width=\textwidth
]

\textbf{\textcolor{blue!70!black}{[GLOBAL CONTEXT]}} \\[2pt]
\begin{tabular}{l|ccc}
\textit{True$\backslash$Pred} & \textbf{0} & \textbf{1} & \textbf{2} \\
\hline
\textbf{0} & 6 & \cellcolor{red!25}\textbf{25} & 7 \\
\textbf{1} & 0 & 9 & 10 \\
\textbf{2} & 0 & 0 & 7 \\
\end{tabular}
\hspace{5pt}
\begin{minipage}[t]{0.55\textwidth}
\textbf{Error Distribution:}\\
$\bullet$ $0 \to 1$: 25 errors (59.5\%) $\leftarrow$ \textit{CURRENT FOCUS}\\
$\bullet$ $1 \to 2$: 10 errors (23.8\%)\\
$\bullet$ $0 \to 2$: 7 errors (16.7\%)
\end{minipage}

\par\noindent\rule{\linewidth}{0.5pt}\par
\vskip 4pt

\textbf{\textcolor{blue!70!black}{[LOCAL ERROR EXAMPLES]}} $\mathcal{E}_{0 \to 1}$ \\[2pt]
\begin{tabular}{|p{0.92\textwidth}|}
\hline
\rowcolor{red!8}
\textbf{Example 1:} ``\textit{They understand it partially because I understand what they are doing but I would show them that setting it up as a proportion makes more sense.}'' \\
\textbf{True:} 0 \quad \textbf{Pred:} 1 \quad \textbf{Reasoning:} Teacher acknowledges some understanding but lacks specific analysis... \\
\hline
\rowcolor{red!8}
\textbf{Example 2:} ``\textit{His explanation doesn't really refer to the equivalent ratios. Although the actual work seems to show he has some understanding.}'' \\
\textbf{True:} 0 \quad \textbf{Pred:} 1 \quad \textbf{Reasoning:} Vague acknowledgment without depth... \\
\hline
\end{tabular}

\par\noindent\rule{\linewidth}{0.5pt}\par
\vskip 4pt

\textbf{\textcolor{blue!70!black}{[CONTRASTIVE CORRECT EXAMPLES]}} $\mathcal{E}^{+}_{0}, \mathcal{E}^{+}_{1}$ \\[2pt]
\begin{tabular}{|p{0.45\textwidth}|p{0.45\textwidth}|}
\hline
\rowcolor{green!8}
\multicolumn{1}{|c|}{\textbf{Correctly Classified as 0}} & \multicolumn{1}{c|}{\textbf{Correctly Classified as 1}} \\
\hline
``\textit{Student A has a limited understanding because he based his answer on the dollar amount.}'' & ``\textit{The student looked at price versus amount to find the unit rate.}'' \\
\scriptsize{$\rightarrow$ Incorrect/generic analysis} & \scriptsize{$\rightarrow$ Procedural description of student work} \\
\hline
\end{tabular}

\par\noindent\rule{\linewidth}{0.5pt}\par
\vskip 4pt

\textbf{\textcolor{blue!70!black}{[TASK]}} Analyze WHY the $0 \to 1$ confusion occurs. Identify misleading patterns and propose rule fixes.

\end{tcolorbox}
\caption{Example Reflector prompt for analyzing the dominant error mode $(0 \to 1)$. The prompt provides: (1) global context via the full confusion matrix, (2) local error examples with model reasoning traces, and (3) contrastive correct examples to establish decision boundaries.}
\label{fig:reflector_prompt}
\end{figure*}

\begin{figure}[t]
\centering
\small
\begin{tcolorbox}[
    colback=blue!3,
    colframe=blue!60!black,
    title={\textbf{Reflector Output: Diagnosis $g_{0 \to 1}$}},
    fonttitle=\bfseries\small,
    boxrule=0.5pt
]

\textbf{\textcolor{red!70!black}{Root Cause:}} \\
The classifier misinterprets teachers' acknowledgment of ``some understanding'' as sufficient for Score 1. Phrases like ``valid reasoning'' or ``understands partially'' mislead the classifier despite lacking analytical depth.

\vskip 4pt

\textbf{\textcolor{red!70!black}{Misleading Patterns:}} \\
$\bullet$ Positive language (``some understanding'', ``valid reasoning'') \\
$\bullet$ Missing specific evidence from student work \\
$\bullet$ Vague acknowledgments without multiplicative analysis

\vskip 4pt

\textbf{\textcolor{red!70!black}{Why These Are 0, Not 1:}} \\
Per rubric: Score 0 for responses that provide \textit{generic} or \textit{absent} analysis. Simply stating ``valid reasoning'' $\neq$ analyzing mathematical thinking.

\vskip 4pt

\textbf{\textcolor{green!60!black}{Proposed Rule Fix:}} \\
\ding{182} Responses acknowledging partial understanding \textbf{must} include specific examples demonstrating multiplicative relationships. \\
\ding{183} Vague acknowledgments without detailed analysis $\Rightarrow$ Score 0.

\vskip 4pt

\textbf{\textcolor{orange!80!black}{Safety Check:}} \\
This fix reinforces depth requirements, which aligns with addressing $1 \to 2$ errors (also require specificity). Will not increase other error modes.

\end{tcolorbox}
\caption{Structured diagnosis output from the Reflector for mode $(0 \to 1)$. The output identifies root causes, misleading patterns, and proposes targeted rule modifications with safety considerations.}
\label{fig:reflector_output}
\end{figure}

\begin{figure}[t]
\centering
\small
\begin{tcolorbox}[
    colback=white,
    colframe=black!70,
    title={\textbf{Refiner Prompt for Rule Generation}},
    fonttitle=\bfseries\small,
    boxrule=0.5pt
]

\textbf{\textcolor{blue!70!black}{[INPUT]}}

\vskip 2pt
$\bullet$ Current rules: Rules$^{(t)}$ \\
$\bullet$ Diagnosis: $g_{0 \to 1}$ (from Reflector) \\
$\bullet$ Error examples: $\mathcal{E}_{0 \to 1}$ \\
$\bullet$ Other modes: $\{(1,2), (0,2)\}$ ← \textit{Cross-mode awareness}

\vskip 4pt
\hrule
\vskip 4pt

\textbf{\textcolor{blue!70!black}{[CONSTRAINTS]}}

\vskip 2pt
\ding{182} Fix must target $0 \to 1$ confusion specifically \\
\ding{183} Must NOT break existing correct classifications \\
\ding{184} Must be COMPATIBLE with fixes for $(1 \to 2)$, $(0 \to 2)$ \\
\ding{185} Edit budget: \texttt{medium} (2--3 new rules)

\vskip 4pt
\hrule
\vskip 4pt

\textbf{\textcolor{blue!70!black}{[OUTPUT FORMAT]}}

\vskip 2pt
Generate improved rules that: \\
$\bullet$ Include clear distinguishing criteria \\
$\bullet$ Use specific patterns from error analysis \\
$\bullet$ State safety considerations for other modes

\end{tcolorbox}
\caption{Refiner prompt structure showing how cross-mode awareness is incorporated. The Refiner receives information about all active error modes to generate compatible rule fixes.}
\label{fig:refiner_prompt}
\end{figure}

\subsection{Confusion-Aware Error Analysis}

The diagnostic phase begins by explicitly evaluating the current rubric $\mathcal{P}^{(t)}$ on the sampled minibatch to generate a set of predictions and their corresponding reasoning traces. By comparing these predictions $\hat{y}$ against ground truth $y$, we construct a confusion matrix $\mathbf{C}^{(t)}$ (lines 4--7). Based on this performance data, we construct a comprehensive context for the optimizer. For each identified confusion mode $(i, j) \in \mathcal{M}$, we assemble three distinct types of information to ground the model's reasoning. First, we extract \textbf{local error context}. This consists of specific examples $\mathcal{E}_{i \to j} = \{(x, y, \hat{y}, r)\}$ where the true label $y$ is $i$, but the model predicted $\hat{y} = j$, accompanied by the model's reasoning trace $r$ (line 12). Second, to establish clear decision boundaries, we sample \textit{contrastive correct examples}. These are instances correctly classified as $i$ or $j$ but which exhibit high model uncertainty (line 13). We quantify this uncertainty using a misconfidence score \cite{chu2025llm}:
\begin{equation}
\text{misconf}(x) = \begin{cases} 
-\log P_\theta(\hat{y}|x) & \text{if } \hat{y} = y \\ 
\left|\frac{\log P_\theta(\hat{y}|x)}{\log P_\theta(y|x)}\right| & \text{if } \hat{y} \neq y 
\end{cases}
\label{eq:misconf}
\end{equation}
where $P_\theta(c|x)$ denotes the predicted probability for class $c$. High misconfidence in correct predictions signals boundary cases that provide high-value discriminative signals. Finally, we provide \textbf{global context} via the full confusion matrix summary, $\text{GlobalContext}(\mathbf{C})$. This summary includes the error distribution across all classes, ensuring that the optimizer remains aware of the global performance landscape and avoids overfitting to a single error type at the expense of others.

\subsection{Mode-Specific Feedback Generation}
To repair the identified errors, CARO avoids the ``rule dilution" of baseline methods by generating separate diagnosis for each error mode (lines 10--16). We employ a Reflector LLM to analyze the confusion pattern $(i, j)$ specifically. The gradient generation process is formalized as:
\begin{equation}
g_{i \to j} = \text{Reflector}(\text{Rules}^{(t)}, \mathcal{E}_{i \to j}, \mathcal{E}^{+}_{i}, \mathcal{E}^{+}_{j}, \text{GlobalContext}(\mathbf{C}), \mathcal{M})
\label{eq:gradient}
\end{equation}
Here, $\mathcal{E}^{+}_{i}$ and $\mathcal{E}^{+}_{j}$ represent the contrastive correct examples. We have shown a demonstration of the Reflector's prompt in Figure \ref{fig:reflector_prompt}.

As exemplified by Figure \ref{fig:reflector_output}, the resulting gradient $g_{i \to j}$ provides a structured diagnosis, identifying the root cause of the confusion, creating discriminative criteria to distinguish score $i$ from $j$, and performing a safety check to predict potential negative side effects on other score categories.

\subsection{Two-Phase Rule Consolidation}
A critical innovation of CARO is the consolidation strategy, which converts discrete error diagnostics into a unified, coherent rubric (lines 18--23). We execute this in two phases to balance specificity with logical consistency. 

\paragraph{Phase 1: Per-Mode Rule Generation} 
We translate the gradients into specific rule refinements (lines 19--22). For each mode $(i, j)$, a Refiner module generates a rule update $r_{i \to j}$. Crucially, the Refiner receives information about all other targeted error modes $\mathcal{M} \setminus \{(i,j)\}$ as additional context. This \textit{cross-mode awareness} enables the Refiner to anticipate potential interactions. We show an example of the prompt to Refiner in Figure \ref{fig:refiner_prompt}: when generating a rule to prevent $0 \to 1$ misclassification, it can avoid phrasing that would inadvertently increase $1 \to 2$ or $0 \to 2$ errors if these modes are also being addressed in the same round.

\paragraph{Phase 2: Priority-Weighted Consolidation} 
Rather than simply concatenating all new rules, which often leads to bloated or contradictory prompts, we organize rules based on the frequency of the error modes they address (line 23). The consolidated rule set for the next iteration is derived as:
\begin{equation}
\text{Rules}^{(t+1)} = \text{Consolidate}(\{r_{i \to j}\}_{(i,j) \in \mathcal{M}}, \mathbf{C})
\label{eq:consolidate}
\end{equation}
As shown in Figure \ref{fig:consolidation}, the consolidation assigns explicit precedence levels: the dominant error mode (highest $C_{ij}$) receives Priority 1 with 2--3 detailed rules; secondary modes receive Priority 2+ with single-sentence guards. When rules from different priority levels suggest different scores for an edge case, we insert explicit \textit{tie-breaker directives}, e.g., ``If Priority 1 criteria suggest score $X$ but Priority 2 criteria suggest score $Y$, assign score $X$ unless [specific exception condition].'' This prioritization ensures that the optimization creates a stable ``diagnosis-and-repair" pipeline, systematically eliminating the most significant failure patterns before addressing long-tail errors.

\begin{figure*}[t]
\centering
\small
\begin{tcolorbox}[
    colback=white,
    colframe=black!70,
    title={\textbf{Two-Phase Rule Consolidation}},
    fonttitle=\bfseries\small,
    boxrule=0.5pt,
    width=\textwidth
]

\begin{minipage}[t]{0.53\textwidth}
\textbf{\textcolor{blue!70!black}{Phase 1: Per-Mode Rules}}

\vskip 4pt
\fcolorbox{red!50}{red!5}{
\begin{minipage}{0.95\textwidth}
\textbf{$r_{0 \to 1}$} (25 errors --- Priority 1)\\[2pt]
\scriptsize
\ding{182} Responses with ``partial understanding'' without specific student work examples $\Rightarrow$ Score 0 \\
\ding{183} Vague acknowledgments lacking multiplicative analysis $\Rightarrow$ Score 0 \\
\ding{184} Must explicitly connect to proportional reasoning
\end{minipage}
}

\vskip 6pt
\fcolorbox{orange!50}{orange!5}{
\begin{minipage}{0.95\textwidth}
\textbf{$r_{1 \to 2}$} (10 errors --- Priority 2)\\[2pt]
\scriptsize
\ding{182} ``Strong understanding'' without specific calculations $\Rightarrow$ Score 1 \\
\ding{183} Must show explicit scaling/proportional evidence for Score 2
\end{minipage}
}

\vskip 6pt
\fcolorbox{yellow!50}{yellow!5}{
\begin{minipage}{0.95\textwidth}
\textbf{$r_{0 \to 2}$} (7 errors --- Priority 3)\\[2pt]
\scriptsize
\ding{182} Generic praise (``good number sense'') without evidence $\Rightarrow$ Score 0
\end{minipage}
}

\end{minipage}
\hfill
\begin{minipage}[t]{0.46\textwidth}
\textbf{\textcolor{blue!70!black}{Phase 2: Consolidated Rules $r_{\text{consolidated}}$}}

\vskip 4pt
\fcolorbox{green!50}{green!5}{
\begin{minipage}{0.95\textwidth}
\textbf{Priority 1: $0 \to 1$ Guard (CHECK FIRST)}\\[2pt]
\scriptsize
$\bullet$ Vague acknowledgment (``some understanding'', ``valid reasoning'') \textbf{without} specific student work $\Rightarrow$ \textbf{Score 0} \\
$\bullet$ Must include specific calculations showing multiplicative relationships

\vskip 4pt
\textbf{Priority 2: Score 1 vs 2 Boundary}\\[2pt]
\scriptsize
$\bullet$ Procedural description only $\Rightarrow$ Score 1 \\
$\bullet$ Explicit proportional reasoning with evidence $\Rightarrow$ Score 2

\vskip 4pt
\textbf{Conflict Resolution:}\\[2pt]
\scriptsize
If Priority 1 criteria suggest Score 0 but response mentions correct calculation, apply Score 0 \textbf{unless} the response explicitly analyzes \textit{why} the calculation demonstrates proportional understanding.
\end{minipage}
}

\end{minipage}

\end{tcolorbox}
\caption{Two-phase rule consolidation. \textbf{Phase 1} generates per-mode rules with cross-mode awareness. \textbf{Phase 2} synthesizes them into a priority-weighted consolidated ruleset with explicit conflict resolution directives.}
\label{fig:consolidation}
\end{figure*}

\subsection{Candidate Selection and Sampling}
The final phase bridges the current iteration to the next by performing two distinct operations (lines 24--26): (1) \textit{candidate selection}, which determines which prompt variants effectively resolved errors and should survive to the next round, and (2) \textit{minibatch sampling}, which constructs the optimal training data for the subsequent iteration to prevent overfitting and ensure robust evaluation. 

\paragraph{Diversity-Aware Candidate Selection} CARO maintains a beam of $B$ candidate prompts across iterations. Each candidate prompt $\mathcal{P}$ generated in ``Two-Phase Rule Consolidation'' is tagged with its target mode $m(\mathcal{P}) \in \mathcal{M}$, indicating which confusion cell it was designed to address. To prevent the beam from collapsing to variants of a single mode fix, we employ a greedy selection strategy that balances empirical performance with mode coverage:
\begin{equation}
\text{SelectScore}(\mathcal{P}) = \widetilde{\kappa}(\mathcal{P}) + \lambda \cdot \mathbb{1}[m(\mathcal{P}) \notin \mathcal{S}]
\label{eq:select}
\end{equation}
where $\widetilde{\kappa}(\mathcal{P})$ is the min-max normalized Cohen's $\kappa$ score \cite{cohen1960coefficient}, $\mathcal{S}$ represents the set of error modes already covered by previously selected candidates in the current selection round, and $\lambda$ is a hyperparameter controlling the diversity bonus. The indicator function $\mathbb{1}[\cdot]$ awards a bonus to candidates targeting modes not yet represented in the selected set $\mathcal{S}$, encouraging exploration of diverse rule modifications.

\paragraph{Semantic Minibatch Sampling} Instead of random minibatch construction, we use an SBERT~\cite{reimers2019sentence}-based similarity selector to focus optimization effort on challenging regions of the input space. We retrieve training examples that are semantically similar to the high-misconfidence instances identified in previous rounds:
\begin{equation}
\text{Minibatch}^{(t)} = \bigcup_{x \in \text{TopMisconf}^{(t-1)}} \text{kNN}_k(x, \mathcal{D}_{\text{train}})
\label{eq:minibatch}
\end{equation}
where $\text{kNN}_k$ retrieves the $k$ nearest neighbors based on SBERT embeddings. This ensures that refined rules are stress-tested against examples where the current rubric is most ambiguous, accelerating convergence on difficult boundary cases.

\subsection{Inference with Optimized Prompt}
Upon the termination of the optimization procedure, we select the rubric prompt $\mathcal{P}^*$ that maximizes the evaluation metric (Cohen's $\kappa$) on the validation set:
\begin{equation}
\mathcal{P}^* = \arg\max_{\mathcal{P} \in \mathcal{P}^{(T)}} \kappa(\mathcal{P}, \mathcal{D}_{\text{val}})
\label{eq:select_best}
\end{equation}
This globally optimized prompt serves as the fixed instruction set for the final inference phase on the held-out test dataset $\mathcal{D}_{\text{test}}$. 
For a given student response $x \in \mathcal{D}_{\text{test}}$, the predicted score is obtained as:
\begin{equation}
\hat{y} = \arg\max_{y \in \mathcal{Y}} P_\theta(y \mid x, \mathcal{P}^*)
\label{eq:inference}
\end{equation}
where $P_\theta(y \mid x, \mathcal{P}^*)$ denotes the probability assigned by the LLM predictor $f_\theta$ to score $y$ given the student response $x$ and the optimized rubric $\mathcal{P}^*$. 
The model generates a completion that includes both a reasoning trace $r$ and a final numerical score, extracted via regular expression parsing:
\begin{equation}
(\hat{y}, r) = \text{Parse}\big(f_\theta(x \oplus \mathcal{P}^*; \tau=0)\big)
\label{eq:parse}
\end{equation}
By enforcing the reasoning trace through the structure of $\mathcal{P}^*$, the inference process provides not only a quantitative grade but also an interpretable justification that reflects the resolved decision boundaries established during optimization.

\section{Experiments}

\subsection{Experimental Setup}

\paragraph{Datasets}
We evaluate CARO on three educational grading datasets spanning K-12 science education and teacher education:

\textbf{Interaction Dataset ($\mathcal{D}_I$):} This dataset contains 252 high school student responses to a physics assessment item on electrical interactions \cite{kaldaras2021developing}. Each response is evaluated against 11 binary rubric categories (labeled 0/1), where 0 indicates the requirement is ``not met'' and 1 indicates it is ``met.'' The categories assess distinct aspects of student understanding, ranging from basic concept recognition to complex reasoning about charge interactions.

\textbf{Elementary Item Response Dataset ($\mathcal{D}_E$):} This dataset comprises elementary student responses to five science assessment tasks graded on an ordinal scale $\{0, 1, 2, 3, 4\}$. The tasks assess foundational science concepts with varying levels of complexity.

\textbf{Teacher Education Dataset ($\mathcal{D}_T$):} This dataset evaluates teachers' pedagogical content knowledge in mathematics education \cite{copur2022mathematics}. It contains eight grading tasks organized into three knowledge dimensions:
\begin{itemize}[noitemsep,topsep=0pt]
    \item \textbf{KSMT} (Knowledge of Students' Mathematical Thinking): Tasks T1--T3 assess teachers' ability to interpret student mathematical reasoning.
    \item \textbf{KMT} (Knowledge of Mathematical Tasks): Tasks T4--T6 evaluate understanding of task design and cognitive demand.
    \item \textbf{KT} (Knowledge of Teaching): Tasks T7--T8 assess pedagogical decision-making.
\end{itemize}
All tasks in $\mathcal{D}_T$ use ordinal labels $\{0, 1, 2\}$, representing absent, partial, and complete understanding respectively.

We partition each dataset into train/validation/test splits using a 7:1:2 ratio following \cite{chu2025llm}. Dataset statistics are summarized in Table~\ref{tab:dataset_stats}.

\begin{table}[ht]
\centering
\caption{A summary of statistics of the datasets}
\label{tab:dataset_stats}
\setlength{\tabcolsep}{2.2pt}
\begin{tabular}{ccccc}
\hline
\textbf{Dataset} & \textbf{Domain} & \textbf{\#Samples} & \textbf{Scores} \\
\hline
$\mathcal{D}_I$ & Interaction & 252 & \{0,1\}  \\
$\mathcal{D}_T$ & Teacher Education & 224$\sim$236 & \{0,1,2\}  \\
$\mathcal{D}_E$ & Elementary & 199$\sim$434 & \{0,1,2,3,4\}  \\
\hline
\end{tabular}
\end{table}

\paragraph{Baselines}
We compare CARO against two baselines.
First is \textbf{Naive Prompting}, which uses direct zero-shot prompting with an expert-written rubric, without any optimization.
Secondly, we compare with \textbf{GradeOpt} \cite{chu2025llm}, a state-of-the-art automatic prompt optimization framework that iteratively refines rubrics using aggregated error feedback.

\paragraph{Implementation Details}
We use \texttt{GPT-4o-mini} as both the grading predictor $f_\theta$ and the optimization agents (Reflector and Refiner) across all methods for fair comparison. For CARO, we set: top-$K$ error modes $K=4$, beam size $B=4$, diversity weight $\lambda=0.3$, and maximum rounds $T=6$. The edit budget is set to ``medium'' (2--3 rule modifications per mode). For GradeOpt, we use the default hyperparameters from \cite{chu2025llm} and set the maximum rounds to $T=6$ to align with computational constraints of CARO. To mitigate the inherent stochasticity of LLM generation and ensure the reliability of our findings, we conducted 3 independent runs for all methods (CARO and baselines) across each task. The reported results represent the average performance on the held-out test set across these 3 trials. 

\paragraph{Evaluation Metrics}
We report classification accuracy (Acc) and Cohen's $\kappa$ coefficient \cite{cohen1960coefficient}. While accuracy measures overall correctness, $\kappa$ accounts for class imbalance and chance agreement, making it the primary metric for grading reliability assessment. A $\kappa > 0.6$ is generally considered substantial agreement \cite{landis1977measurement}.

\subsection{Main Results}

\paragraph{Teacher Education Dataset ($\mathcal{D}_T$)}
Table~\ref{tab:te_results} presents results on the teacher education dataset. CARO achieves consistent improvements across all eight tasks, with an average accuracy improvement of $+11\%$ over GradeOpt and $+41\%$ over Naive prompting.

\begin{table}[ht]
\centering
\caption{Performance Comparison for Teacher Education Dataset $\mathcal{D}_{T}$}
\label{tab:te_results}
\setlength{\tabcolsep}{3pt}
\begin{tabular}{c|c|cc|cc|cc}
\toprule
\hline
\textbf{Type} & \textbf{Task} &
\multicolumn{2}{c|}{\textbf{Naive}} &
\multicolumn{2}{c|}{\textbf{GradeOpt}} &
\multicolumn{2}{c}{\textbf{CARO}} \\
 &  & \textbf{Acc} & \textbf{$\kappa$} & \textbf{Acc} & \textbf{$\kappa$} & \textbf{Acc} & \textbf{$\kappa$} \\
\hline
KSMT & T1  & 0.40 & 0.20 & 0.61 & 0.41 & 0.67 & 0.47 \\
KSMT & T2 & 0.58 & 0.42 & 0.74 & 0.61 & 0.79 & 0.69 \\
KSMT & T3  & 0.39 & 0.09 & 0.51 & 0.25 & 0.55 & 0.30 \\
\textbf{KSMT} & \textbf{Avg} & \textbf{0.46} & \textbf{0.24} & \textbf{0.62} & \textbf{0.42} & \textbf{0.67} & \textbf{0.49} \\
\midrule
KMT  & T4  & 0.30 & 0.05 & 0.55 & 0.28 & 0.76 & 0.57 \\
KMT  & T5  & 0.59 & 0.28 & 0.60 & 0.31 & 0.74 & 0.46 \\
KMT  & T6  & 0.68 & 0.44 & 0.76 & 0.58 & 0.76 & 0.59 \\
\textbf{KMT}  & \textbf{Avg} & \textbf{0.52} & \textbf{0.26} & \textbf{0.64} & \textbf{0.39} & \textbf{0.75} & \textbf{0.54} \\
\midrule
KT   & T7  & 0.45 & 0.29 & 0.74 & 0.60 & 0.77 & 0.65 \\
KT   & T8   & 0.60 & 0.34 & 0.70 & 0.48 & 0.72 & 0.52 \\
\textbf{KT}   & \textbf{Avg} & \textbf{0.52} & \textbf{0.32} & \textbf{0.72} & \textbf{0.54} & \textbf{0.75} & \textbf{0.59} \\
\midrule
\textbf{Overall} & \textbf{Avg} & \textbf{0.51} & \textbf{0.27} & \textbf{0.65} & \textbf{0.44} & \textbf{0.72} & \textbf{0.53} \\
\hline
\bottomrule
\end{tabular}
\end{table}

The most substantial gains occur on tasks with severe baseline confusion patterns. For instance, on T4 (KMT), CARO improves $\kappa$ from 0.28 (GradeOpt) to 0.57, which is a relative improvement of 104\%. This task exhibits significant $0 \to 1$ confusion in the baseline, which CARO directly addresses through mode-specific gradient generation. Similarly, T5 shows a $\kappa$ improvement from 0.31 to 0.46, demonstrating CARO's effectiveness on tasks where GradeOpt's aggregate feedback fails to resolve specific confusion patterns.
Notably, CARO's improvements are most pronounced in the KMT dimension, where grading criteria involve subtle distinctions between ``no understanding'' and ``partial understanding'', which is precisely the type of adjacent-class confusion that CARO is designed to resolve.

\paragraph{Interaction Dataset ($\mathcal{D}_I$)}
Table~\ref{tab:interaction_results} shows results on the binary classification tasks. Despite the simpler label space, several categories exhibit class imbalance that inflates accuracy while depressing $\kappa$.
CARO achieves the highest $\kappa$ on 10 of 11 categories. The improvements are particularly striking on challenging categories:
\begin{itemize}[noitemsep,topsep=0pt]
    \item \textbf{Category 3}: $\kappa$ improves from 0.61 (GradeOpt) to 0.89, a 46\% relative gain.
    \item \textbf{Category 4}: $\kappa$ improves from 0.36 to 0.62, resolving severe confusion in this low-agreement task.
    \item \textbf{Category 8}: Accuracy jumps from 0.44 to 0.74 (+68\%), with $\kappa$ improving from 0.04 to 0.12.
\end{itemize}

Categories 6, 8, and 9 exhibit persistently low $\kappa$ despite high accuracy improvements, indicating extreme class imbalance where the minority class remains difficult to identify. Nevertheless, CARO substantially outperforms baselines in accuracy, suggesting improved discrimination even when $\kappa$ is constrained by distributional factors.
\begin{table}[htbp]
\centering
\caption{Performance Comparison for Interaction Dataset $\mathcal{D}_{I}$}
\label{tab:interaction_results}
\setlength{\tabcolsep}{4pt}
\begin{tabular}{c|cc|cc|cc}
\toprule
\hline
\textbf{Category} &
\multicolumn{2}{c|}{\textbf{Naive}} &
\multicolumn{2}{c|}{\textbf{GradeOpt}} &
\multicolumn{2}{c}{\textbf{CARO}} \\
 & \textbf{Acc} & $\kappa$ & \textbf{Acc} & $\kappa$ & \textbf{Acc} & $\kappa$ \\
\hline
1 & 0.94 & 0.79 & 0.96 & 0.84 & 0.98 & 0.92 \\
2 & 0.86 & 0.71 & 0.88 & 0.76 & 0.93 & 0.87 \\
3 & 0.82 & 0.52 & 0.85 & 0.61 & 0.95 & 0.89 \\
4 & 0.52 & 0.19 & 0.64 & 0.36 & 0.81 & 0.62 \\
5 & 0.62 & 0.18 & 0.95 & 0.72 & 0.96 & 0.78 \\
6 & 0.42 & 0.03 & 0.73 & 0.09 & 0.87 & 0.21 \\
7 & 0.77 & 0.11 & 0.95 & 0.48 & 0.97 & 0.54 \\
8 & 0.22 & 0.03 & 0.44 & 0.04 & 0.74 & 0.12 \\
9 & 0.15 & -0.04 & 0.27 & 0.01 & 0.90 & -0.02 \\
10 & 0.42 & 0.14 & 0.72 & 0.41 & 0.82 & 0.47 \\
11 & 0.60 & 0.14 & 0.90 & 0.50 & 0.94 & 0.64 \\
\midrule
\textbf{Avg} & \textbf{0.58} & \textbf{0.25} & \textbf{0.75} & \textbf{0.43} & \textbf{0.90} & \textbf{0.55} \\
\hline
\bottomrule
\end{tabular}
\end{table}

\begin{table}[ht]
\centering
\caption{Performance Comparison for EIR Dataset $\mathcal{D}_{E}$}
\label{tab:eir_results}
\setlength{\tabcolsep}{4pt}
\begin{tabular}{c|cc|cc|cc}
\toprule
\hline
\textbf{Task} &
\multicolumn{2}{c|}{\textbf{Naive}} &
\multicolumn{2}{c|}{\textbf{GradeOpt}} &
\multicolumn{2}{c}{\textbf{CARE}} \\
 & \textbf{Acc} & \textbf{$\kappa$} & \textbf{Acc} & \textbf{$\kappa$} & \textbf{Acc} & \textbf{$\kappa$} \\
\hline
T1  & 0.16 & -0.08 & 0.63 & 0.35 & 0.63 & 0.39 \\
T2  & 0.49 & 0.30 & 0.74 & 0.64 & 0.75 & 0.66 \\
T3  & 0.57 & 0.25 & 0.79 & 0.67 & 0.81 & 0.70 \\
T4  & 0.48 & 0.20 & 0.71 & 0.57 & 0.74 & 0.62 \\
T5  & 0.50 & 0.04 & 0.70 & 0.46 & 0.73 & 0.56 \\
\midrule
\textbf{Avg} & \textbf{0.44} & \textbf{0.14} & \textbf{0.71} & \textbf{0.54} & \textbf{0.73} & \textbf{0.59} \\
\hline
\bottomrule
\end{tabular}
\end{table}

\paragraph{Elementary Item Response Dataset ($\mathcal{D}_E$)}
Table~\ref{tab:eir_results} presents results on the elementary response dataset. CARO achieves consistent improvements across all five tasks, with an average $\kappa$ of 0.59 compared to 0.54 for GradeOpt and 0.14 for Naive prompting.
The most notable improvement occurs on T5, where $\kappa$ increases from 0.46 (GradeOpt) to 0.56, which represents a 22\% relative improvement. This task involves distinguishing between partial and complete understanding, a boundary that benefits from CARO's contrastive example selection.

\paragraph{Summary of Improvements}
Table~\ref{tab:summary} summarizes the aggregated performance across all datasets. On average, CARO improves accuracy from 0.51 (Naive prompting) to 0.78, representing a 53\% increase compared to GradeOpt (37\% increase from 0.51 to 0.70). CARO achieves an overall average $\kappa$ of 0.56, representing a 19\% relative improvement over GradeOpt (0.47) and a 155\% improvement over Naive prompting (0.22). 

\begin{table}[t]
\centering
\caption{Summary of average performance across datasets.}
\label{tab:summary}
\setlength{\tabcolsep}{4pt}
\begin{tabular}{l|cc|cc|cc}
\toprule
\textbf{Dataset} & \multicolumn{2}{c|}{\textbf{Naive}} & \multicolumn{2}{c|}{\textbf{GradeOpt}} & \multicolumn{2}{c}{\textbf{CARO}} \\
 & Acc & $\kappa$ & Acc & $\kappa$ & Acc & $\kappa$ \\
\midrule
$\mathcal{D}_I$ (avg) & 0.58 & 0.25 & 0.75 & 0.43 & \textbf{0.90} & \textbf{0.55} \\
$\mathcal{D}_T$ (avg) & 0.51 & 0.27 & 0.65 & 0.44 & \textbf{0.72} & \textbf{0.53} \\
$\mathcal{D}_E$ (avg) & 0.44 & 0.14 & 0.71 & 0.54 & \textbf{0.73} & \textbf{0.59} \\
\midrule
\textbf{Overall} & 0.51 & 0.22 & 0.70 & 0.47 & \textbf{0.78} & \textbf{0.56} \\
\bottomrule
\end{tabular}
\end{table}

\subsection{Ablation Studies}

We conduct ablation studies to analyze the key design choices and advantages of CARO.

\paragraph{Per-Mode vs. Consolidated Rules}
As described in Section 3.4, CARO generates two types of candidate rules during the Two-Phase Rule Consolidation: (1) \textbf{per-mode rules} $r_{i \to j}$ that target individual confusion cells, and (2) \textbf{consolidated rules} $r_{\text{consolidated}}$ that synthesize all per-mode rules into a unified, priority-weighted ruleset via Eq.~\ref{eq:consolidate}. Both types are included in the candidate pool (Algorithm~\ref{alg:caro}, line 24), and the diversity-aware selection mechanism (Eq.~\ref{eq:select}) determines which candidates propagate to subsequent rounds. Table~\ref{tab:ablation_mode} reports which candidate type achieves the best validation $\kappa$ across all tasks.

We observe that per-mode rules are selected as the final best prompt in over 90\% of cases across all datasets. This indicates that targeted fixes for specific confusion cells typically outperform holistic rewrites that attempt to address multiple error patterns simultaneously. The consolidated rules, despite being carefully synthesized with priority weighting and conflict resolution directives, tend to dilute the specificity needed for precise decision boundaries. However, consolidated rules are occasionally preferred (6--8\% of cases) in tasks with strongly interacting error modes, where coherent integration prevents the rule conflicts that can arise from independently applying multiple per-mode fixes.

\begin{table}[t]
\centering
\caption{Ablation: Per-mode vs. consolidated rule selection. Values show the percentage of final best prompts using each strategy.}
\label{tab:ablation_mode}
\setlength{\tabcolsep}{4pt}
\begin{tabular}{l|ccc}
\toprule
\textbf{Dataset} & \textbf{Per-Mode} & \textbf{Consolidated} & \textbf{Best $\kappa$} \\
\midrule
$\mathcal{D}_I$ & 93.9\% & 6.1\% & Per-Mode \\
$\mathcal{D}_T$ & 91.7\% & 8.3\% & Per-Mode \\
$\mathcal{D}_E$ & 93.3\% & 6.7\% & Per-Mode \\
\midrule
\textbf{Avg} & 93.0\% & 7.0\% & - \\
\bottomrule
\end{tabular}
\end{table}

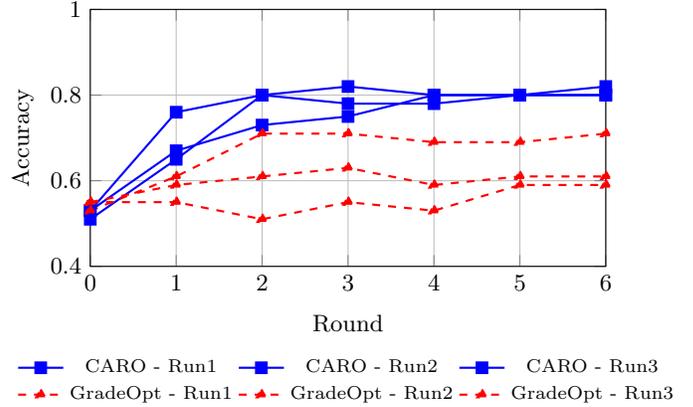
\begin{figure}[t]
\centering
\begin{tikzpicture}
\begin{axis}[
    width=\linewidth,
    height=5cm,
    trim axis left,
    trim axis right,
    xlabel={Round},
    ylabel={Accuracy},
    ylabel near ticks,
    legend style={
        font=\scriptsize,
        at={(0.5,-0.32)},
        anchor=north,
        legend columns=3,
        draw=none
    },
    grid=major,
    xmin=0, xmax=6,
    ymin=0.4, ymax=1.0
]
\addplot[blue, mark=square*, thick] coordinates {(0,0.53) (1,0.76) (2,0.80) (3,0.82) (4,0.80) (5,0.80) (6,0.80)};
\addplot[blue, mark=square*, thick] coordinates {(0,0.53) (1,0.67) (2,0.73) (3,0.75) (4,0.80) (5,0.80) (6,0.80)};
\addplot[blue, mark=square*, thick] coordinates {(0,0.51) (1,0.65) (2,0.80) (3,0.78) (4,0.78) (5,0.80) (6,0.82)};
\addplot[red, mark=triangle*, thick, dashed] coordinates {(0,0.55) (1,0.59) (2,0.61) (3,0.63) (4,0.59) (5,0.61) (6,0.61)};
\addplot[red, mark=triangle*, thick, dashed] coordinates {(0,0.55) (1,0.55) (2,0.51) (3,0.55) (4,0.53) (5,0.59) (6,0.59)};
\addplot[red, mark=triangle*, thick, dashed] coordinates {(0,0.53) (1,0.61) (2,0.71) (3,0.71) (4,0.69) (5,0.69) (6,0.71)};
\legend{CARO - Run1, CARO - Run2, CARO - Run3, GradeOpt - Run1, GradeOpt - Run2, GradeOpt - Run3}
\end{axis}
\end{tikzpicture}
\caption{Convergence comparison on T4 ($\mathcal{D}_I$). CARO achieves faster convergence and higher final accuracy.}
\label{fig:convergence}
\end{figure}

\paragraph{Convergence Analysis}
Figure~\ref{fig:convergence} shows the optimization trajectory across rounds for a representative task T4 in $\mathcal{D}_I$ over three independent runs. Several patterns emerge from this comparison.

\textbf{Rapid initial gains.} CARO demonstrates substantially faster early-stage improvement, achieving an average accuracy of 0.69 after just one round (from 0.52 baseline), compared to GradeOpt's 0.58. By Round 2, all CARO runs reach or exceed 0.73 accuracy, while all GradeOpt runs remain below 0.71.

\textbf{Stability vs. oscillation.} CARO exhibits monotonic or near-monotonic improvement with stable convergence around 0.80 accuracy by Round 3--4. In contrast, GradeOpt displays the oscillation characteristic of aggregate feedback methods: Run 2 drops from 0.55 to 0.51 at Round 2, and Run 1 regresses from 0.63 to 0.59 at Round 4. This instability reflects mode interference, where fixing one error pattern inadvertently amplifies another.

\textbf{Final performance gap.} After 6 rounds, CARO achieves a final accuracy of 0.80--0.82 across all runs (avg. 0.81), while GradeOpt plateaus at 0.59--0.71 (avg. 0.64). This gap demonstrates that CARO's mode-specific updates not only converge faster but also reach superior optima by avoiding the destructive interference inherent in consolidated feedback approaches.

\paragraph{API Cost Analysis}
Table~\ref{tab:cost} compares the computational cost of CARO and GradeOpt. Despite using the same number of optimization rounds, CARO requires 80\% fewer API calls, translating to approximately 60\% cost reduction.
This efficiency benefits from CARO's targeted optimization strategy: rather than exhaustively sampling errors across all confusion cells to generate aggregate feedback, CARO focuses gradient generation on only the top-$K$ error modes per round. Additionally, the mode-specific Reflector and Refiner operate on refined subsets of error examples (Eq.~\ref{eq:gradient}), whereas GradeOpt's aggregate approach requires broader sampling to capture diverse error patterns within a single feedback signal.

\begin{table}[t]
\centering
\caption{API cost comparison (average per task). Costs estimated using GPT-4o-mini pricing (\$0.15/1M input, \$0.60/1M output tokens).}
\label{tab:cost}
\setlength{\tabcolsep}{5pt}
\begin{tabular}{l|ccc}
\toprule
\textbf{Method} & \textbf{Rounds} & \textbf{API Calls} & \textbf{Est. Cost} \\
\midrule
GradeOpt & 6 & $\sim$40,000 & $\sim$\$5 \\
CARO & 6 & $\sim$8,000 & $\sim$\$2 \\
\midrule
\textbf{Reduction} & -- & 80\% & 60\% \\
\bottomrule
\end{tabular}
\end{table}

\begin{figure}[t]
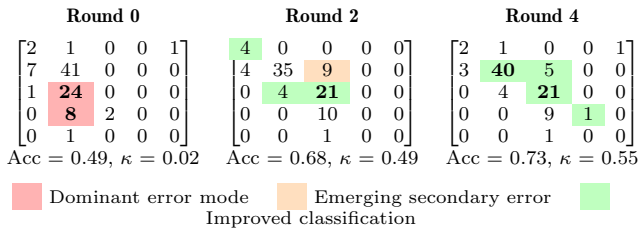

\centering
\small
\resizebox{1.05\columnwidth}{!}{%
\begin{tabular}{ccc}
\textbf{Round 0} & \textbf{Round 2} & \textbf{Round 4} \\[4pt]
$\begin{bmatrix} 
2 & 1 & 0 & 0 & 1 \\ 
7 & 41 & 0 & 0 & 0 \\ 
1 & \cellcolor{red!30}\mathbf{24} & 0 & 0 & 0 \\ 
0 & \cellcolor{red!30}\mathbf{8} & 2 & 0 & 0 \\ 
0 & 1 & 0 & 0 & 0 
\end{bmatrix}$ &
$\begin{bmatrix} 
\cellcolor{green!25}4 & 0 & 0 & 0 & 0 \\ 
4 & 35 & \cellcolor{orange!25}9 & 0 & 0 \\ 
0 & \cellcolor{green!25}4 & \cellcolor{green!25}\mathbf{21} & 0 & 0 \\  
0 & 0 & 10 & 0 & 0 \\ 
0 & 0 & 1 & 0 & 0 
\end{bmatrix}$ &
$\begin{bmatrix} 
2 & 1 & 0 & 0 & 1 \\ 
3 & \cellcolor{green!25}\mathbf{40} & \cellcolor{green!25}5 & 0 & 0 \\ 
0 & 4 & \cellcolor{green!25}\mathbf{21} & 0 & 0 \\  
0 & 0 & 9 & \cellcolor{green!25}1 & 0 \\ 
0 & 0 & 1 & 0 & 0 
\end{bmatrix}$ \\[6pt]
Acc = 0.49, $\kappa$ = 0.02 & Acc = 0.68, $\kappa$ = 0.49 & Acc = 0.73, $\kappa$ = 0.55
\end{tabular}
}
\vskip 4pt
\scriptsize
\colorbox{red!30}{\phantom{X}} Dominant error mode \quad
\colorbox{orange!25}{\phantom{X}} Emerging secondary error \quad
\colorbox{green!25}{\phantom{X}} Improved classification

\caption{Confusion matrix progression on T5 ($\mathcal{D}_E$). Rows represent true labels; columns represent predictions. CARO's iterative refinement first resolves the dominant $2 \to 1$ collapse (Round 0--2), then addresses the emergent $1 \to 2$ over-correction (Round 2--4), achieving balanced diagonal improvement.}
\label{fig:cm_progression}
\end{figure}

\subsection{Qualitative Analysis}

\paragraph{Case Study: Iterative Error Resolution on T5 ($\mathcal{D}_E$)}

We illustrate CARO's ``diagnosis-and-repair'' pipeline through a detailed case study on T5 ($\mathcal{D}_E$), which assesses elementary students' understanding of how a Peregrine Falcon perceives a rock pigeon through a four-step light perception process (light travels to pigeon $\to$ reflects off pigeon $\to$ enters falcon's eyes $\to$ brain processes images). The task uses a 5-point ordinal scale $\{0,1,2,3,4\}$ where Score 2 requires mentioning 1--2 steps, while Score 1 is reserved for responses containing no accurate steps. Figure~\ref{fig:cm_progression} traces the confusion matrix evolution across optimization rounds, revealing how CARO systematically resolves cascading error patterns.

\textbf{Round 0: Identifying the Dominant Mode.} The initial classifier exhibits severe collapse toward Score 1, with the $2 \to 1$ cell containing 24 errors (53\% of all misclassifications, highlighted in red). Analysis reveals that responses mentioning partial steps, such as \textit{``the light source is the sun and the light travels to the eye and that's how the falcon can see the pigeon''}, were incorrectly penalized for omitting steps rather than credited for partial understanding. A secondary $3 \to 1$ pattern (8 errors) compounds this collapse, resulting in zero correct predictions for Scores 2--4 and a near-chance $\kappa = 0.02$.

\textbf{Round 2: Resolving the Primary Collapse.} CARO's Reflector diagnoses the root cause: the original rubric lacks explicit guidance for recognizing partial scientific understanding. The generated rule states: \textit{``If a response includes at least one correct element related to the four steps (e.g., light traveling, reflecting, or entering the eyes), classify as Score 2, provided it does not contain entirely irrelevant information.''} As shown in the Round 2 matrix, this intervention reduces $2 \to 1$ errors from 24 to 4 ($-83\%$, green), while correct Score-2 predictions increase from 0 to 21. Class 0 achieves perfect classification (orange). However, the fix reveals an emergent secondary pattern: $1 \to 2$ confusion (9 errors, orange) as the model now over-awards partial credit to responses lacking any valid steps.

\textbf{Round 4: Addressing the Secondary Mode.} CARO shifts focus to the $1 \to 2$ over-correction, introducing a discriminative rule: \textit{``Responses must include at least one identifiable step describing light's journey; vague statements about `seeing' without mechanistic detail remain Score 1.''} The Round 4 matrix shows this refinement reduces $1 \to 2$ errors from 9 to 5 while improving Score-1 correct predictions to 40 (up from 35). Crucially, the $2 \to 1$ gains are preserved (4 errors), and Score 3 achieves its first correct prediction. The final $\kappa = 0.55$ reflects balanced diagonal improvement across all classes.

This progression exemplifies CARO's core mechanism. Instead of treating error reduction as a single, monolithic task, the framework isolates specific confusion patterns within the matrix. This decomposition turns a complex optimization challenge into a sequential and interpretable repair process. Crucially, this step-by-step approach ensures that corrections for one error type build upon previous improvements without causing regression in other classes.

\section{Discussion}

The experimental results demonstrate that the Confusion-Aware Rubric Optimization (CARO) framework significantly advances the state of automated grading by addressing the structural inefficiencies of prior prompt optimization methods. While existing frameworks such as GradeOpt rely on aggregating heterogeneous error signals, our findings confirm that such aggregation leads to rule dilution, where the optimizer struggles to satisfy conflicting constraints simultaneously. By decomposing the optimization context into distinct error modes via the confusion matrix, CARO transforms the complex prompt refinement into a series of targeted repairs.

A primary implication of this study is the relationship between optimization specificity and efficiency. Our analysis reveals that CARO achieves superior performance with approximately 60\% lower computational costs compared to baseline methods. This efficiency stems from the mode-specific gradient generation, which prevents the optimization process from cycling through redundant rule changes. Instead of requiring extensive exploration to find a marginally better prompt, the structured diagnosis allows the model to identify and resolve dominant failure patterns, such as the confusion between partial and complete understanding, with the first few iterations. This suggests that for ordinal grading tasks, structural error analysis (with confusion matrix) is a more resource-efficient method for improvement than simply increasing the volume of feedback or the size of the optimizer model.

Furthermore, the pedagogical utility of automated assessment relies heavily on interpretability and trust. This distinguishes our approach from traditional automated grading methods, such as latent semantic analysis or BERT-based fine-tuning, which process inputs through high-dimensional vectors and lack visibility into the decision-making process. Unlike these earlier supervised pipelines that function as black-box systems, CARO optimizes the natural language rubric itself. The resulting prompts contain explicit reasoning traces and priority-weighted decision rules. This transparency allows educators to validate the grading criteria before deployment, ensuring that automated system aligns with instructional goals.

Despite these strengths, there are limitations to consider. The current framework relies on the capability of the backbone LLM to self-reflect and generate diagnostic feedback. While our experiments utilized \texttt{GPT-4o-mini} effectively, the performance of CARO on smaller, open-source models remain an area for future investigation.

\section{Conclusion}

This study introduced Confusion-Aware Rubric Optimization (CARO), a novel framework designed to bridge the alignment gap between expert grading criteria and LLM logic. By identifying that grading errors are not random noise but structured clusters, we developed a method that replaces aggregate feedback with mode-specific diagnosis and repair. Through the systematic decomposition of confusion matrices, CARO isolates specific misclassification patterns and generates targeted rule patches that resolve ambiguities without degrading global performance.

Our empirical evaluation across three distinct educational datasets confirms that CARO significantly outperforms state-of-the-art prompt optimization methods in both accuracy and inter-rater reliability. Critically, the framework achieves these gains while reducing computational overhead, offering a scalable solution for high-quality formative assessment. By providing a transparent, efficient, and precise method for rubric refinement, CARO represents a substantial step reliable and scalable automated feedback in complex learning environments.


%
\bibliographystyle{abbrv}
\bibliography{secs/ref}  
\end{document}